\documentclass[10pt,twocolumn,letterpaper]{article}
\usepackage[pagenumbers]{iccv}
\def\confName{ICCV}
\def\confYear{2025}

\def\paperTitle{3D Mesh Editing using Masked LRMs}

\def\authorBlock{
    Will Gao$^{1*}$ \qquad
    \qquad Dilin Wang$^{2}$ \qquad Yuchen Fan$^{2}$ \qquad Aljaz Bozic$^{2}$ \qquad Tuur Stuyck$^{2}$ \\ Zhengqin Li$^{2}$ \qquad Zhao Dong$^{2}$ \qquad Rakesh Ranjan$^{2}$\qquad Nikolaos Sarafianos$^{2}$\\
    $^1$University of Chicago, \qquad $^2$Meta Reality Labs \\\\
    \large{\bf \href{https://chocolatebiscuit.github.io/MaskedLRM/}{MaskedLRM Website}}
}

\newif\ifreview 
\newif\ifarxiv 
\newif\ifcamera 
\newif\ifrebuttal 

\usepackage{graphicx}	
\usepackage{amsmath}	
\usepackage{amssymb}	
\usepackage{booktabs}
\usepackage{times}
\usepackage{microtype}
\usepackage{epsfig}
\usepackage{caption}
\usepackage{float}
\usepackage{placeins}
\usepackage{color, colortbl}
\usepackage{stfloats}
\usepackage{enumitem}
\usepackage{tabularx}
\usepackage{xstring}
\usepackage{multirow}
\usepackage{xspace}
\usepackage{url}
\usepackage{subcaption}
\usepackage{xcolor}
\usepackage[hang,flushmargin]{footmisc}
\usepackage[percent]{overpic}

\ifcamera \usepackage[accsupp]{axessibility} \fi

% ...

\newcommand{\nbf}[1]{{\noindent \textbf{#1.}}}

\ifarxiv  \fi

\newcommand{\R}[1]{{%
    \textbf{%
        \ifstrequal{#1}{1}{\textcolor{red}{R#1}}{%
        \ifstrequal{#1}{2}{\textcolor{blue}{R#1}}{%
        \ifstrequal{#1}{3}{\textcolor{magenta}{R#1}}{%
        \ifstrequal{#1}{4}{\textcolor{teal}{R#1}}{%
                           \textcolor{cyan}{R#1}%
        }}}}%
    }%
}}

\usepackage{xcolor}
\definecolor{cvprblue}{rgb}{0.21,0.49,0.74}

\newcommand{\mytilde}{\raise.17ex\hbox{$\scriptstyle\mathtt{\sim}$}}

\definecolor{iccvblue}{rgb}{0.21,0.49,0.74}
\usepackage[pagebackref,breaklinks,colorlinks,allcolors=iccvblue]{hyperref}

\def\confName{ICCV}
\def\confYear{2025}

\title{\paperTitle}
\author{\authorBlock}

\begin{document}
\maketitle
\begin{abstract}
We present a novel approach to shape editing, building on recent progress in 3D reconstruction from multi-view images. 
We formulate shape editing as a conditional reconstruction problem, where the model must reconstruct the input shape with the exception of a specified 3D region, in which the geometry should be generated from the conditional signal. 
To this end, we train a conditional Large Reconstruction Model (LRM) for masked reconstruction, using multi-view consistent masks rendered from a randomly generated 3D occlusion, and using one clean viewpoint as the conditional signal.
During inference, we manually define a 3D region to edit and provide an edited image from a canonical viewpoint to fill that region. 
We demonstrate that, in just a single forward pass, our method not only preserves the input geometry in the unmasked region through reconstruction capabilities on par with SoTA, but is also expressive enough to perform a variety of mesh edits from a single image guidance that past works struggle with, while being \(2-10\times\) faster than the top-performing prior work.
% \blfootnote{*This work was conducted during an internship at Meta}
\end{abstract}    
% \vspace{-0.2cm}
\section{Introduction}
\label{sec:intro}

Automated 3D content generation has been at the forefront of computer vision and graphics research, due to applications in various visual mediums like games, animation, simulation, and more recently, virtual and augmented reality. As research on neural methods for content generation has progressed, there has been significant progress in modifying and applying well-studied 2D methods into the 3D domain.

Recent developments in 3D content generation initially followed a similar path to 2D content generation. Operating in 3D voxel space instead of pixel space, models like VAEs \cite{meng2019vv, renat2024learning, girdhar2016learning, mescheder2019occupancy} and GANs \cite{zheng2022sdfstylegan, chen2019learning} were built, and trained on small-scale datasets~\cite{brock2016generative}. 
These works often demonstrated limited editing capabilities through simple latent operations on their learned representations. Efforts have been made to extend generative diffusion to 3D~\cite{liumeshdiffusion, zhou20213d, vahdat2022lion, li2023diffusion}.
There has also been work done in generative autoregressive models for 3D, which tokenize 3D data in a unique way \cite{nash2020polygen, tang2024edgerunner, sun2020pointgrow, nichol2022point}. 
Furthermore, neural representation techniques such as NeRFs~\cite{mildenhall2021nerf} and Gaussian splatting \cite{kerbl20233d} have introduced an entirely new paradigm for 3D content generation.

Despite significant progress in 3D content generation from scratch, research in editing the shape of \emph{existing} 3D models is underdeveloped. 
Image editing methods benefit from a nearly endless source of data from scraping the internet, while 3D assets typically require a higher level of expertise and specialized tools to create and thus are scarce in comparison. 
The difference in scale is staggering, with the largest image datasets containing billions of samples~\cite{schuhmann2022laion} while the largest 3D datasets contain only millions~\cite{deitke2024objaverse}.
\begin{figure}[t]
    \newcommand{\pl}{-1}
    \begin{overpic}[width=\columnwidth]{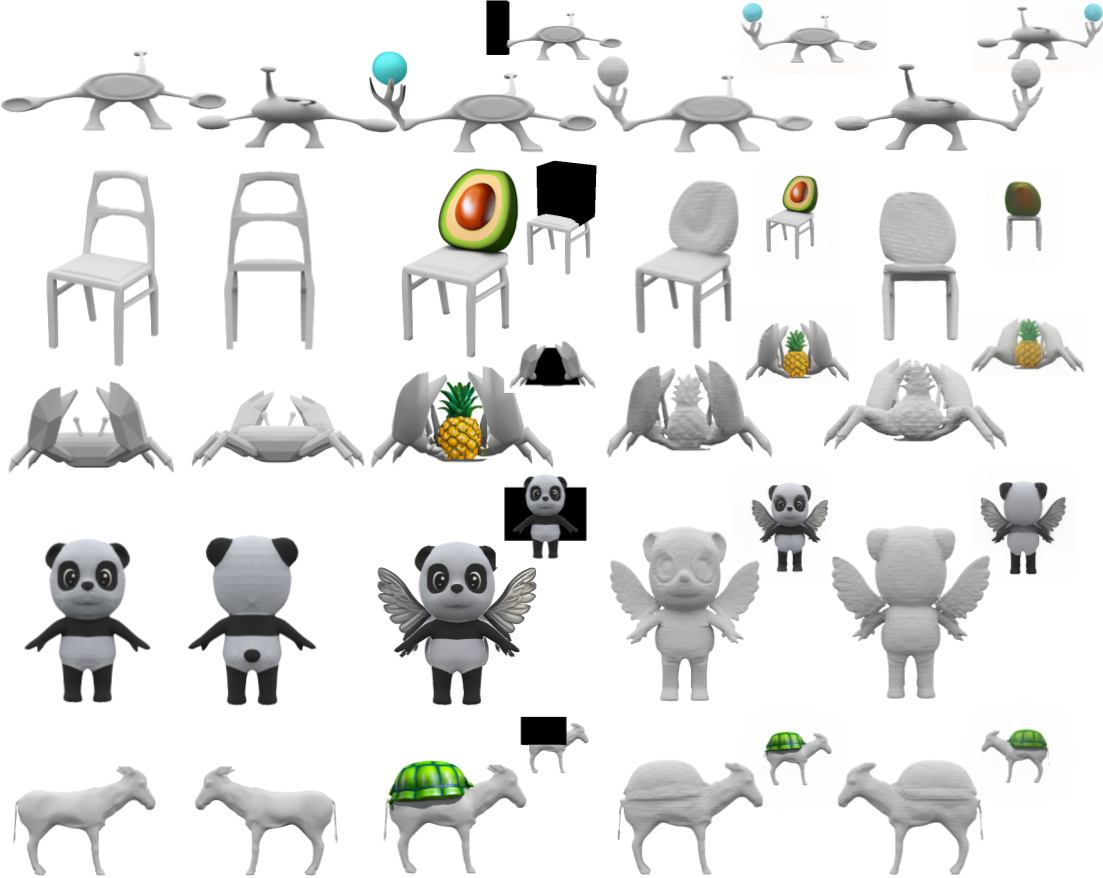} 
    \put(7.5,  -3.5){\textcolor{black}{\textbf{Source}}}
    \put(31.5,  -3.5){\textcolor{black}{\textbf{Mask \& Edit}}}
    \put(61.5,  -3.5){\textcolor{black}{\textbf{LRM Output}}}  
    \end{overpic}
    \vspace{0.05cm}
    \caption{\textbf{Mesh Editing using MaskedLRMs}: The inputs comprise front/back views of the source mesh \((1^{st}\) column) and a frontal image used as the conditional view. 
    The \(2^{nd}\) column shows the masked area rendered from the front (inset) and a 2D edit. 
    The last column shows our generated mesh from the front/back views. 
    % (Inset: colors from volumetric rendering).
    }
\label{fig:qual_examples}
\vspace{-0.2cm}
\end{figure}
A common approach to tackling the issue of 3D data scarcity is to exploit existing text-image foundation models. Recent efforts in 3D editing involve using these huge models to provide guidance to an optimization process by giving them differentiably rendered images of the manipulated geometry as input \cite{michel2022text2mesh, mohammad2022clip, Gao_2023_SIGGRAPH, magicclay_Barda_2024}. 
While these approaches demonstrated some success, they face several major challenges. Firstly, the gradients obtained  using foundation models as guidance are often extremely noisy, leading to unstable and unpredictable optimizations~\cite{wang2023steindreamer}. 
Furthermore, since these methods often use text as input in lieu of visual input, they are hard to control. Finally, these techniques typically directly optimize properties of an explicit 3D mesh, which severely constrains the type of possible edits. For example, it is impossible to add a hole to a shape, since such a modification is not topology-preserving.

Recent works follow a different path and utilize a two-stage approach, placing the brunt of the ``creative effort" onto 2D models, using them to generate and edit content. 
Then, a pipeline that lifts 2D images into 3D content produces the final output \cite{li2023instant3d, xu2024instantmesh}. Thus, by giving the model edited image inputs, a 3D edit is obtained. 
However, these methods rely on diffusion models that produce multi-view images~\cite{liu2023zero, shi2023zero123plus, long2023wonder3d, liu2023syncdreamer} which then are passed to a 3D reconstruction model~\cite{wei2024meshlrm, hong2023lrm}. While editing a single image is no longer a challenging task, this multi-view generation procedure often suffers from ambiguous 3D structure in the occluded regions and does not accurately reconstruct what a shape looks like from every viewpoint. Efforts have been made to adapt multi-view diffusion models specifically for text-driven editing instead of 
the single-view-to-multi-view task~\cite{barda2024instant3ditmultiviewinpaintingfast, PrEditor3D}. As we qualitatively demonstrate, editing multi-view images in a realistic manner remains a challenging task.

Our proposed approach falls into the second direction: lifting 2D images to 3D. 
Instead of using a 3D model to simply reconstruct geometry, our model is inherently trained to ``inpaint" regions of multi-view images. The inpainting task is performed directly in 3D, instead of in multi-view image space.
Specifically, the inputs to our method are a set of masked renders and a single clean conditional image that is provided to infer the missing information from. 
Our approach solves the issues present in both approaches to shape editing. 
In contrast to optimization methods, our model is efficient as it constructs shapes in a single, fast forward pass. 
Furthermore, the output of our model is highly predictable, as it is trained to reconstruct and inpaint geometry to a high degree of accuracy. This predictability gives a high degree of control to our method via the conditioning image. 
Our approach addresses the multi-view consistency and ambiguity problems of reconstruction methods by relying on a single conditional image while propagating the conditional signal to the rest of the multi-view inputs.

A key challenge is designing a training procedure that allows the model to learn how to use the conditional information in a multi-view consistent manner. 
To accomplish this, we introduce a new 3D masking strategy. 
We mask each input view in a consistent manner by rendering an explicit occluding mesh. Then, by supervising both the occluded and unoccluded regions with multi-view reconstruction targets, our model learns to not only fill in the occluded region, but also to accurately reconstruct the rest of the shape. 
Unlike previous works such as NeRFiller~\cite{weber2023nerfiller} which used fixed masks at a per-scene basis, training with randomly generated masks allows our model to generalize to arbitrary shapes and test-time masks. 
We demonstrate that this training method allows our model to be used downstream for editing tasks while maintaining strong quantitative performance on reconstruction baselines. 
By manually defining an editing region analogous to the train-time occlusions, and using a single edited canonical view, users can use our model to generate a shape that is faithful both to the original shape, and the edited content.
In summary, our contributions are as follows:
\begin{itemize}
    \item We design a novel conditional LRM trained with a new 3D-consistent multi-view masking strategy that enables our LRM to generalize to \emph{arbitrary} masks during inference.
    \item Despite not being our primary intention, our architecture matches SoTA reconstruction metrics, while concurrently learns to use the conditional input to fill 3D occlusions. 
    \item We show that our LRM can be used for 3D shape editing while being \(2-10\times\) faster than optimization- and LRM-based edit methods. It synthesizes  edits that optimization cannot (\eg genus changes) and does not suffer from the multi-view consistency and occlusion ambiguity issues that approaches trained without masking suffer from.
\end{itemize}

\section{Related Work}\label{sec:related}
\noindent\textbf{Large Reconstruction Models}: LRM~\cite{hong2023lrm} and its recently introduced variants \cite{li2023instant3d,xu2023dmv3d,wang2023pf,wei2024meshlrm,zhang2025gs,xu2024instantmesh,openlrm,tochilkin2024triposr,sf3d2024} showcase the solid capabilities of the transformer architecture for sparse reconstruction. 
Trained on large-scale 3D~\cite{deitke2023objaverse,deitke2024objaverse} and multi-view image datasets~\cite{yu2023mvimgnet}, these models reconstruct geometry and texture details from  sparse inputs or a single image in a feed-forward manner.
Most LRMs focus on reconstructing radiance fields, which cannot be consumed by standard graphics pipelines for editing. MeshLRM~\cite{wei2024meshlrm} and InstantMesh~\cite{xu2024instantmesh} extract mesh and texture maps, but it remains a challenging to perform shape editing in an intuitive and fast manner. 
Furthermore, while these models achieve quite high reconstruction quality when given at least four complete views as input~\cite{li2023instant3d, zhang2025gs}, the problem is much more ambiguous when given only a single (possibly edited) image~\cite{hong2023lrm}.
In this work we investigate how to utilize the LRM representation power for 3D shape editing, given a handful incomplete views as input for the shape reconstruction. 
This makes the reconstructed geometry of the non-edited content match significantly better to the original geometry, while ensuring view-consistency of the edited parts.
\begin{figure*}[t]
    \centering
    \includegraphics[width=0.92\linewidth]{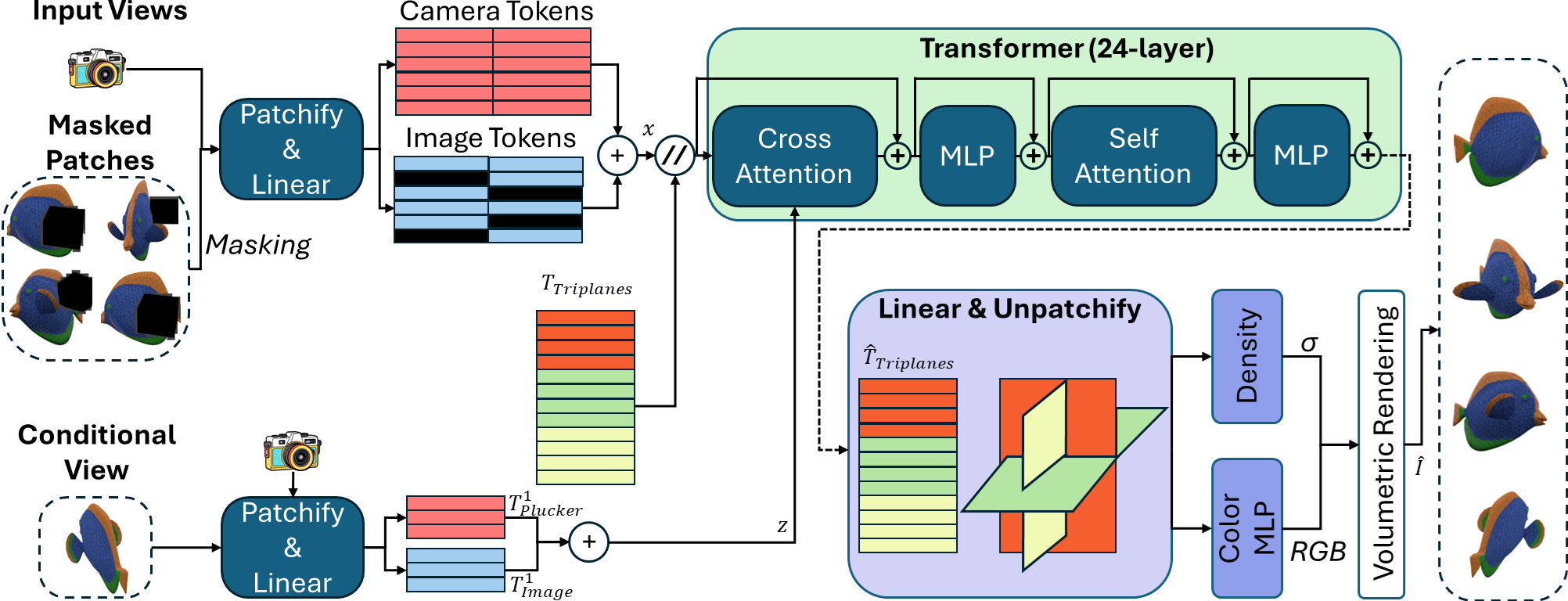}
\caption{\textbf{Training Pipeline}. The images and camera poses are patchified and projected into tokens. 
A random 3D mask is generated and tokens corresponding to occluded patches are replaced by a learnable mask token. 
Camera and image tokens are summed and concatenated with learnable triplane tokens to form the transformer input. 
A clean conditional image is tokenized, forming the cross-attention input. The output triplane tokens are upsampled and decoded into colors and SDF values, which are transformed into densities for volumetric rendering. 
% \ab{Density MLP to Distance MLP to be consistent with text (or SDF to $\sigma$} (Plücker rays) 
}
\label{fig:overview}
\vspace{-0.3cm}
\end{figure*}

\noindent\textbf{Shape Editing}:
Editing 3D shapes has been an active area of research for at least four decades. 
Early works focused on deformation~\cite{sederberg1986free, coquillart1990extended}, cut and paste~\cite{ranta1993cut, biermann2002cut}, Laplacian surface editing~\cite{gal2009iwires, lipman2004differential, laplacianEditing} or fusion~\cite{kanai1999interactive}. 
Recent works have tackled this task from different viewpoints depending on the geometry representation, the losses, and the downstream application. 
Regarding representation, research has been conducted on implicit surface editing~\cite{hertz2022spaghetti, neumesh}, voxels~\cite{sella2023voxe}, mesh-based deformations~\cite{Gao_2023_SIGGRAPH, meshup_Kim_2024, sarafianos2024garment3dgen}, NeRF~\cite{instructnerf2023, weber2023nerfiller, yuan2022nerf, neuraleditor, bao2023sine, NeRFshop23, jung2024geometry} and Gaussian splatting~\cite{wang2024view, chen2023gaussianeditor, kotovenko2025wast, rai2025uvgs}. 
Another line of work focused on generative local editing using diffusion models. 
MagicClay~\cite{magicclay_Barda_2024} sculpted 3D meshes using a 2D hybrid representation where part of the geometry is frozen and the remaining is optimized using SDS loss~\cite{poole2022dreamfusion, li2023focaldreamer}. 
3D shape editing has been explored in the context of sketches, ~\cite{takayama2013sketch, nealen2005sketch, SketchDream2024}, faces~\cite{aneja2023clipface, han2023headsculpt, Fruehstueck2023VIVE3D, chaudhuri2021semi, potamias2024shapefusion} or in an interactive manner~\cite{dong2024coin3d}.
Recent approaches build upon progress in LRMs~\cite{hong2023lrm, wei2024meshlrm}, performing multi-view editing using diffusion models and then using LRMs to reconstruct an edited shape~\cite{barda2024instant3ditmultiviewinpaintingfast, qi2024tailor3dcustomized3dassets, PrEditor3D}. 
In contrast, our work introduces a novel architecture trained on multi-view consistent masked data that bypasses the need for inconsistent diffusion editing and enables 3D mesh editing within seconds. 

\noindent\textbf{Masked Vision Models}:  
The original Denoising Autoencoder~\cite{vincent2008extracting} used random noise as a ``masking" method on images during training with the objective of learning better semantic representations. 
More recently, methods using transformers convert images into sequences and predict unknown pixels~\cite{baobeit, chen2020generative} which culminated in the development of the Vision Transformer (ViT)~\cite{dosovitskiy2020image} as the backbone of modern masked image models. 
Models like the Masked Autoencoder~\cite{he2022masked} use ViTs to process randomly masked tokens, where every token represents a distinct, non-overlapping image patch
Research in diffusion models which also uses random noise as a ``masking" procedure, has exploded in popularity, producing increasingly impressive generated images. 
By taking random Gaussian noise and constraining it to a specific region, diffusion models can be used for image inpainting~\cite{levin2023differential, couairondiffedit}. 
Masked autoencoders have been built for 3D data types such as point clouds \cite{pang2022masked, zhang2022point, jiang2023masked}, meshes \cite{liang2022meshmae}, and NeRFs \cite{irshad2025nerf}, with each work developing a different way to ``patchify" their respective 3D representations. 
Point clouds have the most natural representation for next token prediction \cite{sun2020pointgrow, nichol2022point}, while efforts have also been made into tokenizing triangle meshes for generation as sequences of vertices and faces \cite{nash2020polygen, tang2024edgerunner}.  
Our paper presents a new approach to combine masking with LRMs for editing.

\vspace{-0.2cm}
\section{Method}\label{sec:method}
Our large reconstruction model, shown in Figure~\ref{fig:overview},
reconstructs a 3D shape from input images. 
Specifically, the model maps a sparse set of renders from various viewpoints into a latent triplane representation. 
We sample this representation to obtain latent features at different 3D points, which are then decoded into distance and RGB values for volumetric rendering. 
At training, we predict output renders from arbitrary camera poses. During inference, we use marching cubes to produce the reconstructed geometry.
Unlike existing LRMs, our model uses a conditional branch to accept an additional view of the target shape. The inputs are then corrupted by a random masking procedure, forcing the model to learn to ``inpaint" the input views using the conditional branch signal.

\vspace{-0.15cm}
\subsection{Masked LRM: Architecture}
\label{sec:method:arch}

\noindent\textbf{Image and Pose Tokens.} The raw input to our model is a set of images with known camera parameters. 
During both training and inference, the input shapes are oriented in an arbitrary manner. Since we cannot guarantee a canonical viewpoint in the data, we remove the dependence on absolute poses by computing all camera parameters relative to the first randomly selected image which we use as the conditional input.
These camera parameters are represented by Plücker rays, forming a 6-channel grid with the same spatial dimensions as RGB pixels. 
We apply the standard ViT tokenization~\cite{dosovitskiy2020image} to the image and the Plücker rays independently, dividing both grids into non-overlapping patches, and linearly projecting them into a high-dimensional embedding.

\noindent\textbf{Masking.} After the input images are tokenized, we randomly select a subset of tokens to mask out. For general masked image modeling, \cite{he2022masked} demonstrated that dropping out random patches from the encoded image enabled a desirable balance between reconstruction and learned representation quality. However, since our goal is to train a model that fills in the missing geometry from the content of a single clean view, it is not suitable to occlude random patches since they lack correspondence for each input view. Instead, we require a structured, \emph{3D-consistent} form of occlusion.
Specifically, we generate a 3D rectangular mesh with uniformly random side lengths. We then render the depth map of this mesh from the same cameras as the input images, obtaining a set of multi-view consistent occlusions. Patches containing pixels that would be occluded by this random mesh are masked out. Instead of dropping the masked patches entirely as in~\cite{he2022masked}, we propose to replace them with a learnable token. This does not suffer the same train-test gap, as occluded images are passed to the model during inference as well. This allows the model to maintain the 3D spatial context of the occlusion. Hence, our masking strategy is specifically designed with downstream editing of an occluded shape in mind.

\noindent\textbf{Model Formulation.} Using the above input tokenization and masking procedures, we can write a complete description of our model. 
Let $\mathcal{S}$ be a shape rendered from $n$ camera poses described by the Plücker ray coordinates $\{\mathbf{C_i}\}_{i=1}^n$ producing RGB renders $\{\mathbf{I_i}\}_{i=1}^n$. The input token sequence to our model for any image are given by:
\setlength{\abovedisplayskip}{3pt}
\setlength{\belowdisplayskip}{3pt}
\begin{equation}
    T_\text{Image}^i=\textbf{PatchEmbed}(\mathbf{I}_i),
    T_\text{Plücker}^i=\textbf{PatchEmbed}(\mathbf{C}_i),
\end{equation}
where \textbf{PatchEmbed} is the operation of splitting images into non-overlapping patches and applying a linear layer to the channels.
We reserve $T_\text{Image}^1$ and $T_\text{Plücker}^1$ for the clean conditional signal. Now, we sample a random rectangular mesh $\mathcal{O}$ and render it from the same camera poses as $\mathcal{S}$. Comparing the depth maps of $\mathcal{S}$ and $\mathcal{O}$, we  produce modified tokens $\tilde{T}_\text{Image}^i$ where the token for any patch that contains an occluded pixel is replaced by a learnable mask token. Then, the input tokens to the model are constructed as: 
\begin{equation}
    \mathbf{x} = \underset{i=2} {\overset{n} { \Big \|}} (\tilde{T}_\text{Image}^i + T_\text{Plücker}^i), \; \; \; 
    \mathbf{z} = T_\text{Image}^1 + T_\text{Plücker}^1,
\end{equation}
where $z$ is the condition passed to the model and $\|$ the iterated concatenation operation. 
We choose to add the Plücker ray tokens after masking such that the model can differentiate between different occluded patches. 
Note that adding a Plücker ray token for each patch means the model does not need a positional embedding to differentiate patches.
We use three sequences of learnable tokens $\mathbf{T}_\text{Triplanes} = \mathbf{T}_{xy} || \mathbf{T}_{yz} || \mathbf{T}_{xz}$ to produce the predicted triplanes. 
These tokens are passed to the transformer body of the model, which comprises of iterated standard cross-attention and self-attention operations equipped with MLPs and residual connections:
\begin{equation}
    \mathbf{\hat{x}}, \mathbf{\hat{T}}_\text{Triplanes} = \textbf{Self-Att}\left(\textbf{Cross-Att}\left(\mathbf{x} || \mathbf{T}_\text{Triplanes}\right), \mathbf{z}\right),
\end{equation}
with $\mathbf{x}$ and $\mathbf{T}_\text{Triplanes}$ coming from the previous transformer block (or the input). Finally, we upsample each triplane token to a patch using a single layer MLP, evaluate the learned triplanes at densely sampled points, and decode the latents using MLPs. We obtain predicted images and $\hat{\mathbf{I}}$ pixel-ray opacity maps $\hat{\mathbf{M}}$ through volumetric rendering, and normal maps $\hat{\mathbf{N}}$ by estimating normalized SDF gradients: % \ab{are normal maps also volume rendered?}:
\begin{align*}
    \textbf{Triplanes} &= \textbf{MLP}_\text{Upsample}(\mathbf{\hat{T}_\text{Triplanes}}) \\
    \textbf{SDF}(x, y, z) &= \textbf{MLP}_\text{Distance}(\textbf{Triplanes}(x, y, z)) \\
    \textbf{RGB}(x, y, z) &= \textbf{MLP}_\text{Color}(\textbf{Triplanes}(x, y, z)) \\
    \sigma &= \textbf{Density}(\textbf{SDF}) \\
    \hat{\mathbf{N}} &= \textbf{NormGrad}(\textbf{SDF}) \\
    \hat{\mathbf{I}}, \hat{\mathbf{M}} &= \textbf{VolRender}(\sigma, \textbf{RGB})
\end{align*}
where we convert the SDF values to densities $\sigma$ for rendering following~\cite{yariv2021volume}. The learned image tokens $\mathbf{\hat{x}}$ are not used for any remaining task and are thus discarded.
\vspace{-0.15cm}
\subsection{Supervision}\label{sec:method:losses}
Our LRM is trained with L2 reconstruction and LPIPS perceptual losses. Given a ground truth image $\mathbf{I}$, normal map $\mathbf{N}$, and binary silhouette mask $\mathbf{M}$, we use the following losses:
$$
    \mathcal{L}_\text{Recon} = w_I\|\mathbf{\hat{I}} - \mathbf{I}\|_2^2 + w_N \|\mathbf{\hat{N}} - \mathbf{N}\|_2^2 + w_M\|\mathbf{\hat{M}} - \mathbf{M}\|_2^2
$$
$$
    \mathcal{L}_\text{Percep} = w_P\mathcal{L}_\text{LPIPS}(\mathbf{\hat{I}}, \mathbf{I})
$$
where $w_I, w_N, w_M, w_P$ are tunable weights, and $\mathcal{L}_\text{LPIPS}$ is the image perceptual similarity loss proposed in~\cite{zhang2018perceptual}. For the results in this paper, we choose simply $w_I = w_M = w_P = 1$ and $w_N = 0$ or $1$ depending on the stage of training.
\vspace{-0.1cm}
\subsection{Training Stages}
Our model is trained in stages following~\cite{wei2024meshlrm}, for training efficiency. However, the purpose of our stages differ. Since fully rendering $512\times 512$ output images is computationally expensive, for every stage, we sample a random $128\times 128$ crop from each output image to use as supervision. We maintain the full images for the input throughout every stage.

\noindent\textbf{Stage 1}: We downsample the output images to a $256\times 256$ resolution, allowing the random crops to supervise $25\%$ of the image. We use $128$ samples per ray for the volumetric rendering. 
In this initial stage, we observe that the geometric supervision from the normal maps is not yet necessary, so we drop this portion of the reconstruction loss by setting $w_N=0$ enabling a more efficient backwards pass.

\noindent\textbf{Stage 2}: We downsample the output images to $384\times 384$, meaning that the random crops now only supervise $11\%$ of the image and increase the samples per ray to $512$. 
By increasing the rendering fidelity and decreasing the proportion of the image supervision, we train the model to focus more sharply on geometric details. We observed that without any geometric supervision, the LRM may produce degenerate solutions by generating textures that hide geometric artifacts. Thus, we introduce the normal loss by setting $w_N=1$.

\begin{figure}[t]
\includegraphics[width=\columnwidth]{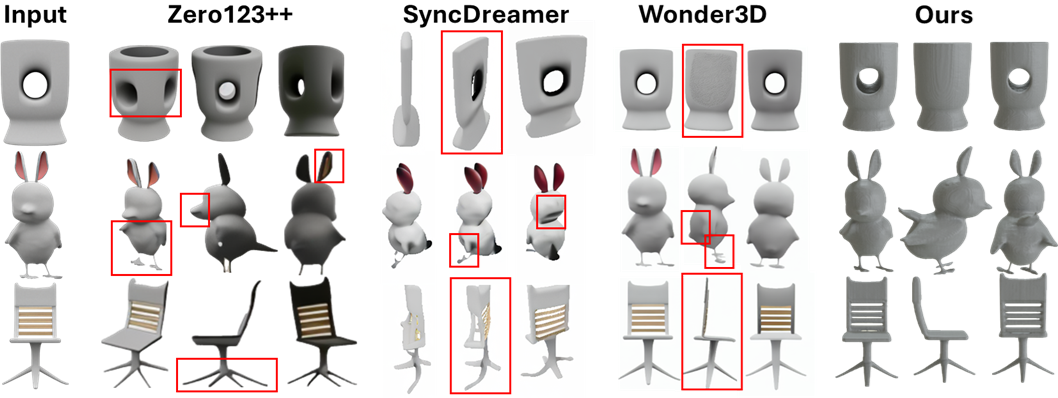}
\caption{\textbf{Multi-view Diffusion vs MaskedLRM}: Multi-view diffusion models must infer occluded geometry from the input which leads to artifacts in the multi-view images (incorrect/distorted views). 
MaskedLRM which does not use multi-view diffusion, receives both the edited image and multi-view information, allowing it to bypass this problem and reconstruct the correct geometry.}
\label{fig:z123}
\vspace{-0.25cm}
\end{figure}
\vspace{-0.15cm}
\subsection{Mesh Shape Editing}
\label{sec:method:editing}
Since our LRM is trained with 3D-consistent, multi-view occlusions, using the conditional branch to complete the partial observations, it is straightforward to use it for shape editing. 
Given a shape $\mathcal{S}$, we manually define an occlusion $\mathcal{O}$ that occludes the region of interest for editing. 
Then, we edit a representative image within the pixels that are occluded from its camera viewpoint. This may be done a variety of ways -- for our results, we use a text-to-image masked diffusion method \cite{levin2023differential, couairondiffedit}. 
The image edit is used as a conditional signal, while the rest of the occluded images are fed to the main transformer body of the LRM. 
The LRM is trained to inpaint the occluded region using the content from the conditional branch, and as such it propagates the 2D edit into 3D space.
This approach to shape editing is much faster than optimization-based methods (see Table~\ref{tab:runtime}), requiring only a single forward pass to lift the desired edit into 3D. 
Our model also produces more realistic shapes since it is trained on a large collection of scanned objects instead of relying on diffusion model guidance and complex regularizations. 
It is also more expressive than optimization-based methods as it can generate arbitrary geometry based on the input condition. For example, it can change the geometric genus of a shape (adding a handle or a hole as in Figure~\ref{fig:topo_examples}), which deformation-based optimization methods cannot do as genus changes are not differentiable.
Generative methods using LRMs such as InstantMesh~\cite{xu2024instantmesh} rely on methods such as Zero-123++~\cite{shi2023zero123plus} to generate multi-view images, introducing view-consistency artifacts. 
In Figure~\ref{fig:z123} we show examples of such artifacts generated by recent models. Zero-123++ hallucinates additional holes in the vase, and generates a distorted and incorrect bird anatomy. SyncDreamer~\cite{liu2023syncdreamer} generates unrealistically distorted views, such as a completely flattened vase, poor bird anatomy, and a warped chair. 
Wonder3D~\cite{long2023wonder3d} is better, but it cannot capture the correct bird anatomy and chair structure. 
In contrast, our model requires only a single view as conditioning and uses the prior from the dataset to construct the shape in a consistent manner. Some recent concurrent work tackles editing directly in the multi-view image space. While this also handles ambiguity, we show in Figure~\ref{fig:qual_baseline1} that our method produces more realistic edits.
% \red{Should we discuss masked diffusion methods like PrEditor3D/Instant3Dit here?}

\vspace{-0.15cm}
\section{Experiments}\label{sec:experiments}

\noindent\textbf{Training Data.} We train our Masked LRM on a 
% 480k 
% subset of
% 
the Objaverse dataset \cite{deitke2023objaverse} containing shapes collected from a wide variety of sources. 
% This subset filters out relatively poor quality 3D models\footnote{Sketchfab data were filtered out due their licence}.
Each shape is normalized to fit within a sphere of fixed radius. Our training data consists of 40 $512\times 512$ images of each shape, rendered from randomly sampled cameras. We also render the corresponding depth and normal maps for these camera poses. Every iteration, the model inputs and reconstruction targets are chosen randomly from these pre-fixed sets of images.

\noindent\textbf{Evaluation.} We evaluate the reconstruction quality of our model on the
GSO \cite{downs2022google} and ABO \cite{collins2022abo} datasets
and compare the state-of-the-art MeshLRM \cite{wei2024meshlrm}. Since MeshLRM cannot be easily repurposed for our editing task, we also compare the reconstruction quality with InstantMesh.
We use PSNR, SSIM, and LPIPS on the output renders from novel poses as metrics. 
To remain consistent with the training setting, we randomly generate a rectangular mask to occlude the input views and provide a different clean view as conditioning for our method.
Finally, we qualitatively demonstrate the main contribution of our model, the ability to propagate 2D edits from a single viewpoint into 3D. We compare our results to prior works for text and image based 3D generation. 

\vspace{-0.1cm}
\subsection{Quantitative Comparisons}\label{sec:experiments:quant}
Table~\ref{tab:baseline_quant} shows novel-view synthesis metrics of our method when compared to InstantMesh and MeshLRM. Since our main goal is to edit existing shapes and not to completely generate shapes from scratch, we choose to train our model by randomly selecting 6-8 input views along with one conditional view, giving our LRM a denser input than MeshLRM. 
We show metrics for both 6 and 8 input views and we compute those on another set of 10 camera poses, different from the input poses.
Our method is competitive with the state-of-the-art model on reconstruction, achieving a 2.56 PSNR improvement
on the ABO dataset, and a comparable PSNR on GSO. 
We observe the same phenomenon in perceptual quality measured by LPIPS, where our method significantly outperforms on ABO shapes, and is comparable on GSO shapes. 
As expected, using 6 views under-performs using 8 views, but only by a slight margin. Furthermore, our method significantly outperforms InstantMesh. This is to be expected, as InstantMesh infers everything from a single view, while both MeshLRM and MaskedLRM access multi-view information.
Our model achieves performance on par with SoTA on reconstructing a diverse set of output poses, indicating that it has learned to effectively ``in-paint" the randomly occluded regions in the input views using context from the available unoccluded signal. 
Since our end goal is mesh shape editing, it is \emph{not critical that we surpass the reconstruction quality} of prior works, as we only need to ensure a high quality baseline for the output geometry. 
We further demonstrate qualitatively in Sec.~\ref{sec:experiments:qual} that our model indeed learns to inpaint using the conditional signal, instead of only the context from multi-view images, thereby accomplishing feed-forward shape editing through a single view.

% \begin{table}[t]
% \caption{\textbf{Quantitative Evaluation}: We evaluate our model using shapes outside of its training set and compare it to the state-of-the-art LRM. 
% We do so using the ABO and GSO shape datasets, 
% % Objaverse dataset
% reconstructing the meshes from 6 and 8 posed images. Although our main objective is not direct reconstruction of new shapes, and our masking procedure introduces additional difficulty in the task, we still achieve better than SoTA metrics
% % }
% on the ABO dataset, and comparable to SoTA metrics on the GSO dataset.}
% \begin{subtable}{\linewidth}
% \centering
% \caption*{Reconstruction metrics on ABO dataset}
% \begin{tabular}{lccc} 
% \toprule
% & PSNR $\uparrow$  & SSIM $\uparrow$ & LPIPS $\downarrow$\\
% \toprule
% MeshLRM~\cite{wei2024meshlrm} & 26.09 & 0.898 & 0.102\\
% Ours (6 views) & 28.37 & 0.946 & 0.081 \\
% Ours (8 views) & \bf 28.65 & \bf 0.947 & \bf 0.078 \\
% \bottomrule
% \end{tabular}
% \end{subtable}
% \;
% \begin{subtable}{\linewidth}
% \centering
% \caption*{Reconstruction metrics on GSO dataset}
% \begin{tabular}{lccc} 
% \toprule
% & PSNR $\uparrow$  & SSIM $\uparrow$ & LPIPS $\downarrow$\\
% \toprule
% MeshLRM~\cite{wei2024meshlrm} & \bf 27.93 & 0.925 & \bf 0.081 \\
% Ours (6 views) & 27.24 & 0.931 & 0.088 \\
% Ours (8 views) & 27.58 & \bf 0.933 & 0.085 \\
% \bottomrule
% \end{tabular}
% \end{subtable}
% \vspace{-0.15cm}
% \label{tab:baseline_quant}
% \end{table}
% \setlength{\belowcaptionskip}{0.01cm}
\begin{table}[t]
\centering
\setlength{\tabcolsep}{0.75mm}
\renewcommand{\arraystretch}{1.2}
\caption{\textbf{Quantitative Evaluation}: We evaluate our model using test-set shapes and compare it to the state-of-the-art LRM and InstantMesh on 
the ABO and GSO shape datasets, 
% Objaverse dataset
reconstructing the meshes from 6 and 8 posed images. 
Despite direct reconstruction of new shapes not being our main goal, and our masking introducing extra difficulty in the task, we still achieve better than SoTA metrics
on ABO, and comparable to SoTA metrics on the GSO dataset.}
\resizebox{\columnwidth}{!}{
\begin{tabular}{lcccccc} 
\toprule
 \multirow{2}{*}{Method} & \multicolumn{3}{c}{ABO Dataset} &  \multicolumn{3}{c}{GSO dataset}\\
 \cmidrule(l){2-4} \cmidrule(l){5-7} 
& PSNR $\uparrow$  & SSIM $\uparrow$ & LPIPS $\downarrow$ & PSNR $\uparrow$  & SSIM $\uparrow$ & LPIPS $\downarrow$\\
\toprule
InstantMesh~\cite{xu2024instantmesh} (NeRF) & - & - & - & 23.14 & 0.898 & 0.119 \\
InstantMesh~\cite{xu2024instantmesh} (Mesh) & - & - & - & 22.79 & 0.897 & 0.120 \\
MeshLRM~\cite{wei2024meshlrm} & 26.09 & 0.898 & 0.102 & \bf 27.93 & 0.925 & \bf 0.081 \\
Ours (6 views) & 28.37 & 0.946 & 0.081 & 27.24 & 0.931 & 0.088 \\
Ours (8 views) & \bf 28.65 & \bf 0.947 & \bf 0.078 & 27.58 & \bf 0.933 & 0.085 \\
\bottomrule
\end{tabular}}
\vspace{-0.25cm}
\label{tab:baseline_quant}
\end{table}

\vspace{-0.1cm}
\subsection{Qualitative Evaluations}\label{sec:experiments:comparisons}
Using a bird mesh generated from a text-to-3D model, and several editing targets, we compare our method against other 3D editing methods. Full results are shown in Figure~\ref{fig:qual_baseline}. 
We define a masked region to edit on the head of the bird (omitted from the figure for brevity). Conditional signals provided to our method generated by masked diffusion are shown in the \(1^{st}\) row, and our results are in the last row.

\nbf{Optimization methods}
In the \(2^{nd}\) and \(3^{rd}\) rows of Figure~\ref{fig:qual_baseline}, we show the results of two text-based mesh optimization methods. Instead of using the edited images themselves, we use the text prompts that we passed to the diffusion model as guidance. 
The first optimization-based method we compare to (\(2^{nd}\) row) is TextDeformer~\cite{Gao_2023_SIGGRAPH}  which uses a Jacobian representation~\cite{njf_Aigerman_2022} for deformations to constrain the edits instead of explicit localization.TextDeformer struggles with the highly localized nature of our task and globally distorts the mesh, failing to produce an output of acceptable quality in all examples. 
We also compare with MagicClay~\cite{magicclay_Barda_2024}, which optimizes both an SDF and a triangle-mesh representation. It also optionally uses a manual ``seed" edit so that the optimization task is easier. 
However, since this requires an additional layer of modeling expertise (\ie a manual user intervention) our method does not require, we opt out of this step for our comparison. 
Unlike TextDeformer, MagicClay selects a subset of vertices to deform to combat noisy SDS~\cite{poole2022dreamfusion} gradients. 
Since we have a 3D mask, we simply choose the vertices that lie within that region.  Although this selection serves to localize the editing process, we observe that the deformations are still noisy. 
While sometimes MagicClay edits are semantically correct (flower and rabbit ears), in other cases such as the fedora (\(3^{rd}\) column) and top hat (\(6^{th}\) column), the optimization process collapses completely.
In both cases, noisy gradients from text-guidance result in results in optimizations that are both unpredictable and uncontrollable.
In contrast, the output of our LRM is highly predictable from the selected conditional view, which may be re-generated until desirable or even manually edited.
\setlength{\abovecaptionskip}{9pt}
\begin{figure}[t]
    \newcommand{\pl}{-1}
    \begin{overpic}[width=0.95\columnwidth]{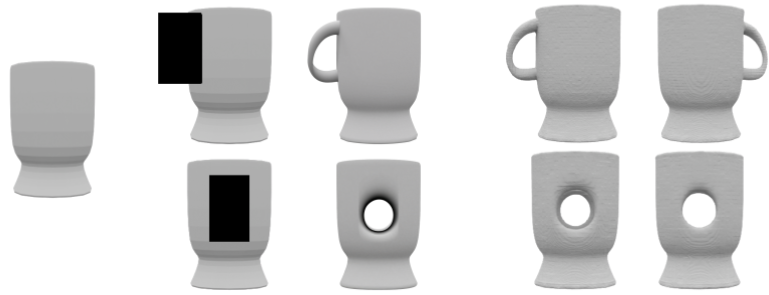} 
    \put(2,  -3.5){\textcolor{black}{\textbf{Source}}}
    \put(26,  -3.5){\textcolor{black}{\textbf{Mask \& Edit}}}
    \put(71,  -3.5){\textcolor{black}{\textbf{LRM Output}}}  
    \end{overpic}
    \vspace{0.02cm}
    \caption{\textbf{Genus changes}: Our method unlocks genus-changing edits like adding a handle or a hole to the original vase. We show the output of our model from 2 opposing views in the \(3^{rd}\) column. 
    % We omit the opposing view from the first column since the source mesh is rotationally symmetric.
    }
\label{fig:topo_examples}
\vspace{-0.3cm}
\end{figure}

\nbf{LRM-based Methods} We compare our method against recent top-performing methods that combine multi-view diffusion and reconstruction models. 
InstantMesh~\cite{xu2024instantmesh} is one such pipeline that may be used for shape editing. It relies on using Zero-123++~\cite{shi2023zero123plus} to generate multi-view images from a single view and then passing these images to an LRM. To edit, we simply pass an edited image of the original shape As shown in the \(4^{th}\) row of Figure~\ref{fig:qual_baseline}, this results in a poorly reconstructed shape that is particularly thin when compared to the ground truth and the output quality suffers due to the inability of Zero-123++~\cite{shi2023zero123plus} to generate faithful multi-view images, as discussed in Section~\ref{sec:method:editing}. 
Methods that rely directly on a separate diffusion module to generate the LRM inputs will run the risk of generating artifacts from inaccurate multiview generation. In comparison, our method does not suffer any such reconstruction artifacts since it uses trivially consistent ground truth renders as the main LRM inputs.

\begin{figure*}[t]
\centering
\includegraphics[width=0.89\linewidth]{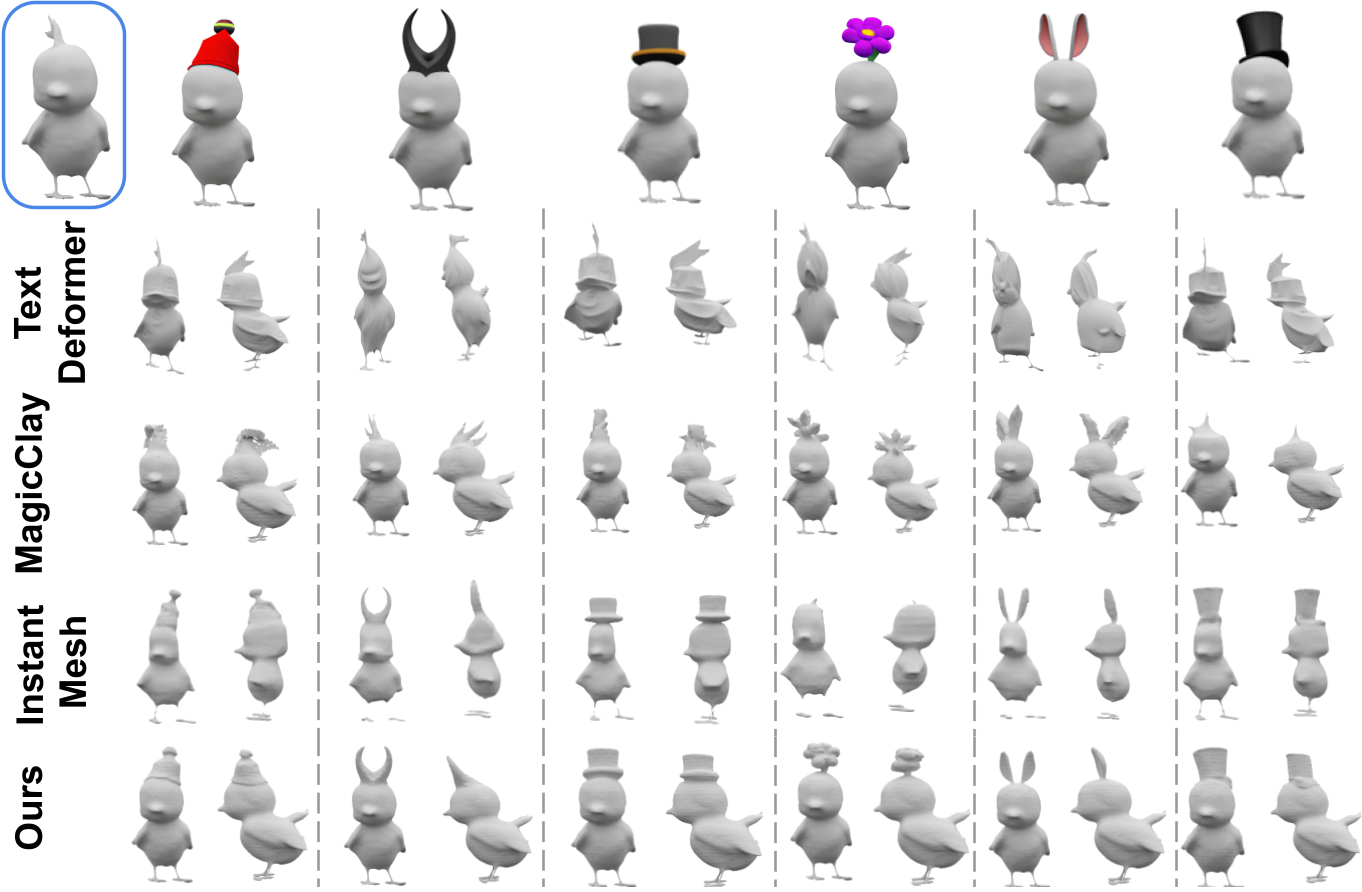}
\caption{\textbf{Mesh Editing Comparisons}: Given a mesh (top left) and various image edits as guidance we demonstrate that our approach is the only one that generates multi-view consistent shape edits that follow the guidance. Colors omitted to clearly visualize the edited geometry.}
%We compare our approach to other shape editing methods. 
\label{fig:qual_baseline}
\vspace{-0.25cm}
\end{figure*}

\begin{figure*}[t]
\centering
\includegraphics[width=0.96\linewidth]{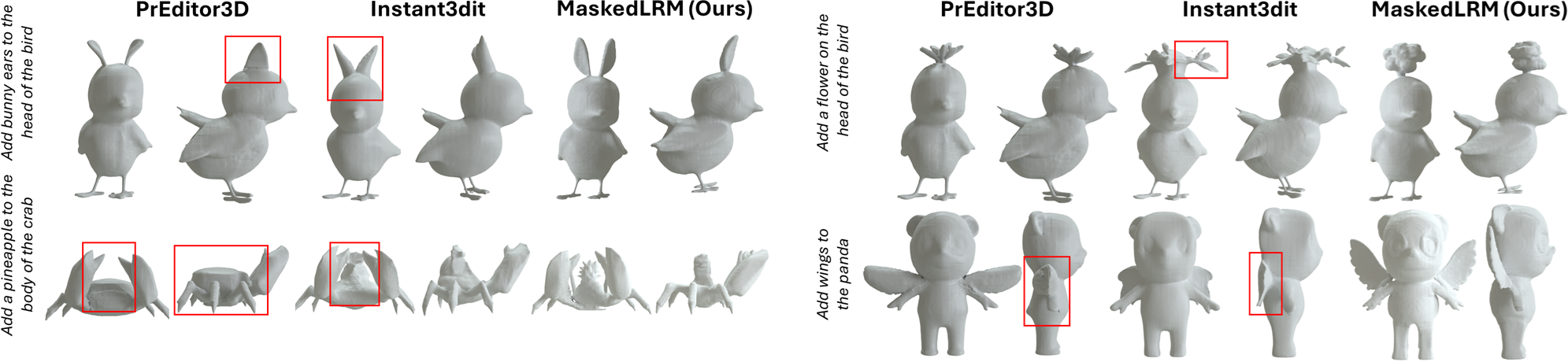}
\caption{\textbf{Mesh Editing Comparisons with Concurrent work}: We compare our approach to concurrent work enabling localized 3D editing. While localized approaches better preserve the original structure of the shape, other methods are not able to produce edits as realistic as ours.}
\label{fig:qual_baseline1}
\vspace{-0.2cm}
\end{figure*}

In Figure~\ref{fig:qual_baseline1} we also compare to two concurrent works namely PrEditor3D~\cite{PrEditor3D} and Instant3Dit~\cite{barda2024instant3ditmultiviewinpaintingfast} that tackle localized editing using multi-view diffusion via text prompting.
% \footnote{While the code is not yet available, the authors of these works offered to run their method on examples we provided which enabled these comparisons.}.
PrEditor3D~\cite{PrEditor3D} performs multi-view editing by first inverting a multi-view diffusion model to obtain Gaussian noise images, and then using the forward process to edit the desired region with a separate text prompt. 
% To produce sharper edits, they use pre-trained models to detect the edited region and merge it with the original shape. 
While PrEditor3D generates semantically correct edits based on the prompts, it produces undesirable artifacts in several examples. In particular, many of the edits lack detail, such as the bunny ears in the top left and the wings in the bottom right. It also fails to produce a pineapple body in the bottom left.
Instant3Dit~\cite{barda2024instant3ditmultiviewinpaintingfast} uses masking in multi-view diffusion training instead of LRM training. 
They use their text-prompted multi-view inpainting model to edit and then use an LRM generate an edited mesh. 
Similar to PrEditor3D, it produces semantically correct edits that lack realism due to artifacts. 
Instead of producing sharp bunny ears, Instant3Dit is only capable of adding vague pointed structures to the bird. In the second column, the flower it generates is plant-like but unrealistic. 
In the bottom row, we see that the pineapple and wings are again semantically correct but lacking in detail.

%
% The second version of InstantMesh, shown in the \(5^{th}\) row, applies the masking procedure from Differential Diffusion~\cite{levin2023differential} to the forward pass of Zero-123++~\cite{shi2023zero123plus}. Once again, since we define a editing region in 3D, we can restrict the diffusion process to the pixels corresponding to this region. In this version, we observe that the reconstruction is higher quality. However, artifacts still remain, especially near the edited region, due to artifacts generated by Zero-123++.

\vspace{-0.15cm}
\subsection{Mesh Editing Characteristics}\label{sec:experiments:qual}
In Figures~\ref{fig:qual_examples} and~\ref{fig:topo_examples}, we show mesh editing examples demonstrating the capabilities of our method. 
The \(1^{st}\) column shows the source mesh rendered from different viewpoints. 
The \(2^{nd}\) column shows the edited conditional image with a render of the original masked region inset.
Our LRM accepts the edited view along with a set of occluded ground-truth renders (omitted from the figure) and predicts an SDF. 
The last column shows the mesh extracted from the output while the insets depict volumetric renders of the predicted SDF.

\nbf{Expressiveness} The edits throughout this paper show the expressiveness of our method. 
The meshes used in Figures~\ref{fig:qual_baseline} as well as in row 1 of Figure~\ref{fig:qual_examples} are examples of non-standard shapes -- a unique bird mesh generated from a text-to-multiview diffusion model~\cite{tang2024mvdiffusionpp} and a ``Tele-alien" from the COSEG~\cite{wang2012active} dataset. 
Despite being novel shapes, our model is able to give the alien a staff and the bird a hat. 
The other four rows of~\ref{fig:qual_examples} consist of edits that are ``unnatural" -- creating an avocado backrest, replacing the body of a crab with a pineapple, giving a panda wings, and giving a donkey a turtle shell. 
In every example, our method successfully translates the 2D edit into geometry in a realistic manner.
The edits in Figure~\ref{fig:topo_examples} show a critical benefit of our method. Since the final mesh is constructed entirely from the output of a neural network, there are no geometric restrictions on the type of edit we can do. The last two rows demonstrate the ability of our network to change the genus of a shape by adding a handle or a hole through the middle, which would be impossible for geometry optimization-based methods.

\nbf{Identity Preservation} Although our model discards the initial shape in order to bypass editing limitations, we observe that the LRM still achieves highly faithful reconstructions of the initial geometry outside of the region of interest. This confirms our quantitative observations that our method has near-SoTA reconstruction quality. This also indicates that, due to multi-view masking, our method is able to constrain the edit inside selected 3D region without needing to perform expensive learning over explicit 3D signals.

\begin{figure}[t]
\includegraphics[width=0.93\columnwidth]{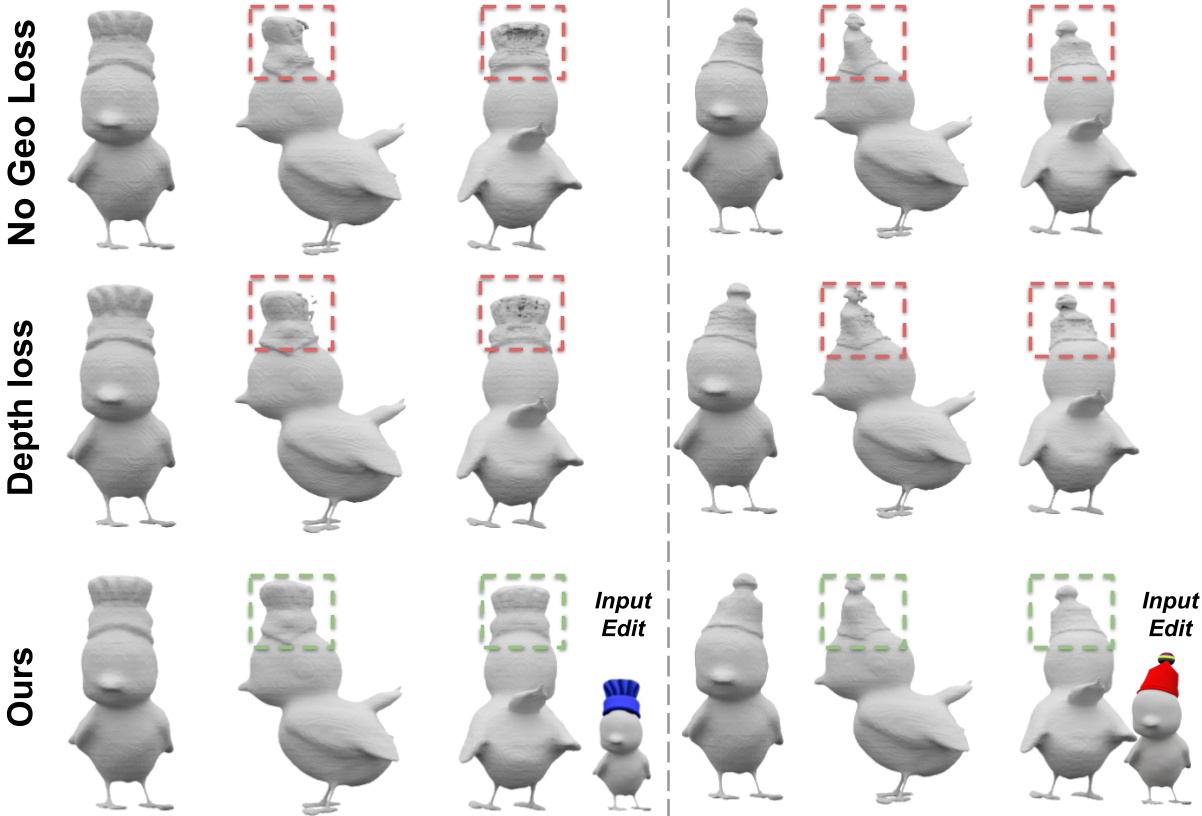}
\caption{\textbf{Impact of Geometric Supervision}: Geometric losses are critical to produce high-quality surfaces. No geometric loss (top) causes severe hole and bump artifacts in the mesh. Depth loss (middle) is not as effective as normal loss (bottom) which allows our model to generate accurate and smooth reconstructions.}
\label{fig:ablation}
\vspace{-0.3cm}
\end{figure}
\vspace{-0.15cm}
\subsection{Ablations Studies \& Discussion}
\nbf{Geometric Supervision} We investigate the effect of using geometric losses during training. Figure~\ref{fig:ablation} compares three LRMs: a model trained by the pipeline described in Sec.~\ref{sec:method}, a model trained with no geometric supervision, and a model trained by replacing the normal map loss with a depth map loss. 
Using no geometric supervision results in poor surface quality, highlighted in the red boxes. Since the main training objective is multi-view image reconstruction, the model hallucinates correct geometry using colors, without producing an accurate surface. 
Supervising the predicted depth somewhat mitigates this issue, but the effect is weak and the surfaces are still incomplete. Normal map supervision gives high quality surfaces as shown in the green boxes.

\nbf{Random Masking} We validate our choice in masking strategy by ablating our method using a uniformly random MAE-style mask across all views. This produces a clear train-test gap as during inference, we are always interested in editing contiguous 3D regions. This gap manifests in blurry and incorrect edits. We refer to Section B and Figure 2 of the supplementary material for details.

\begin{table}[t]
\centering
\setlength{\tabcolsep}{0.75mm}
\renewcommand{\arraystretch}{1.2}
\caption{\textbf{Runtime Comparison}: 
Our method is significantly faster than optimization methods as it is feed-forward and also faster than LRM-based approaches that must run multi-view diffusion.} \label{tab:runtime}
\resizebox{\columnwidth}{!}{
\begin{tabular}{lcccccc} 
\toprule
& \multicolumn{2}{c}{Optimization-based} & \multicolumn{4}{c}{LRM-based}\\
\cmidrule(l){2-3}\cmidrule(l){4-7}
& TextDeformer & MagicClay & InstantMesh & Instant3dit & PrEditor3D & \textbf{Ours} \\ 
Runtime $\downarrow$ & \mytilde20mins & \mytilde1hour & 30sec & \mytilde6sec & 80sec &  \bf \textless\  3sec \\
\bottomrule
\end{tabular}}
\vspace{-0.4cm}
\end{table}

% \begin{table}[t]
% \centering
% \setlength{\tabcolsep}{0.75mm}
% % \renewcommand{\arraystretch}{1.2}
% \caption{\textbf{Runtime Comparison}: 
% Our method is significantly faster than optimization methods as it is feed-forward and also faster than LRM-based approaches that must run multi-view diffusion.} \label{tab:runtime}
% % \resizebox{\columnwidth}{!}{
% \begin{tabular}{lcr} 
% \toprule
% \textbf{Method} & \textbf{Type} & \textbf{Runtime $\downarrow$ } \\
% \midrule
% TextDeformer & Optim. & 20mins\\
% MagicClay & Optim. & 60mins\\
% \midrule
% InstantMesh & LRM & 30sec\\
% Instant3dit & LRM & 9sec\\
% PrEditor3D  & LRM & 80sec\\
% Ours        & LRM & 3sec\\
% % & \multicolumn{2}{c}{Optimization-based} & \multicolumn{3}{c}{LRM-based}\\
% % \cmidrule(l){2-3}\cmidrule(l){4-6}
% % & TextDeformer & MagicClay & InstantMesh & Instant3dit & \textbf{Ours} \\ 
% % Runtime $\downarrow$ & \mytilde20mins & \mytilde1hour & 30sec & \mytilde9sec & \bf \textless\  3sec \\
% \bottomrule
% \end{tabular}
% % }
% \vspace{-0.25cm}
% \end{table}

\noindent\textbf{Runtime}: In Table~\ref{tab:runtime} we provide performance comparisons between our approach and several top-performing recent works. 
Our method is not only much faster than optimization-based approaches~\cite{magicclay_Barda_2024, Gao_2023_SIGGRAPH} as it requires only one forward pass, but it also outperforms LRM approaches~\cite{xu2024instantmesh, barda2024instant3ditmultiviewinpaintingfast} that make use of a multi-view generation model. 
PrEditor3D~\cite{PrEditor3D} requires the forward pass of several large pre-trained models~\cite{ravi2024sam2segmentimages,liu2024groundingdinomarryingdino}
% (a segmentation model~\cite{ravi2024sam2segmentimages} and Grounded DINO~\cite{liu2024groundingdinomarryingdino}), 
resulting in a longer runtime.

\noindent\textbf{Limitations \& Future Work}: Our method is constrained by the expressiveness of editing in the canonical view. 
While text-to-image models can create a wide range of results, capturing a specific idea may require significant iteration. 
Our method is upper-bounded by the uniformity of the Marching Cubes triangulation, and the LRM reconstruction quality which makes performing edits that require extremely intricate details challenging. 
Blurry artifacts may arise when trying to reconstruct fine details (\eg face) but we did not see such issues with shapes like chairs.
MagicClay~\cite{magicclay_Barda_2024}, manually freezes the un-edited geometry part but we designed a solution without such interventions.
% Furthermore, while our model accurately reconstructs geometry, it replaces the triangulation with one derived from Marching Cubes, which may be undesirable. 
% Finally, our model is limited by the reconstruction quality of modern LRMs which makes performing edits that require extremely intricate details challenging. 
Future work could focus on improving localization by developing techniques to merge the existing triangulation with the edited output.

\vspace{-0.1cm}
\section{Conclusion} \label{sec:conclusion}
\vspace{-0.15cm}
In this paper we introduced a new method to perform 3D shape editing. Our work builds upon the recent progress of LRMs by introducing a novel multi-view input masking strategy during training. 
Our LRM is trained to ``inpaint" the masked region using a clean conditional viewpoint to reconstruct the missing information. During inference, a user may pass a single edited image as the conditional input, prompting our model to edit the existing shape in just one forward pass. We believe our method is an significant advancement in shape editing, allowing users to create accurate and controllable edits without 3D modeling expertise.

{
    \small
    \bibliographystyle{ieeenat_fullname}
    \bibliography{References}
}
\newpage
\maketitlesupplementary
\section*{Introduction}
We refer the interested reader to the supplementary video where we provide a plethora of qualitative results of our method. 
In the following sections we: i) conduct an ablation study that showcases the impact of our masking strategy, ii) showcase qualitative results of 2 recent methods (Nerfiller and Tailor3D) and explain some of their shortcomings, iii) provide implementation details of our method and iv) provide several figures with qualitative results. 

\begin{figure}[t]
\includegraphics[width=0.98\columnwidth]{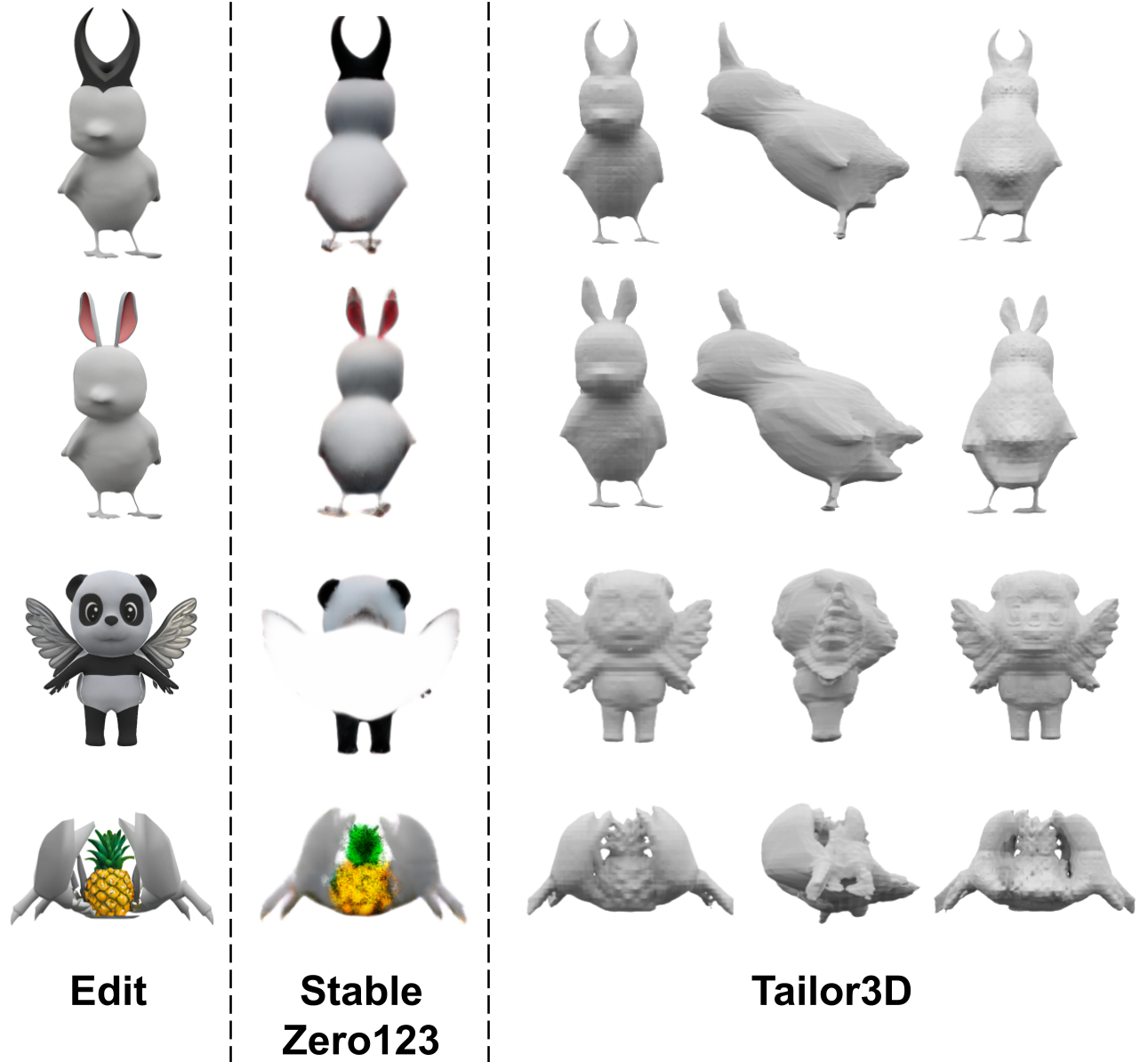}
    \caption{\textbf{Tailor3D Meshes}: Tailor3D results with the some of the conditions we used for our method throughout the paper. The left column shows the source image. The middle column shows a back view generated by Stable Zero-123~\cite{liu2023zero}. The right section shows the Tailor3D geometry rendered from 3 viewpoints. In the first two rows, Tailor3D suffers from ambiguity since it only sees the front and the back views and reconstructs an incorrectly elongated body. In the third row and fourth rows, Stable Zero-123 fails to generate a high-quality back view, failing completely for the wings on the panda. We observe the Janus effect in the generated panda and a lack of sharp features in the generated crab, especially viewed from the side.}
\label{fig:supp:tailor3d}
\vspace{-0.2cm}
\end{figure}

\section*{Masking Ablation}
\begin{figure*}[h]
\includegraphics[width=0.98\textwidth]{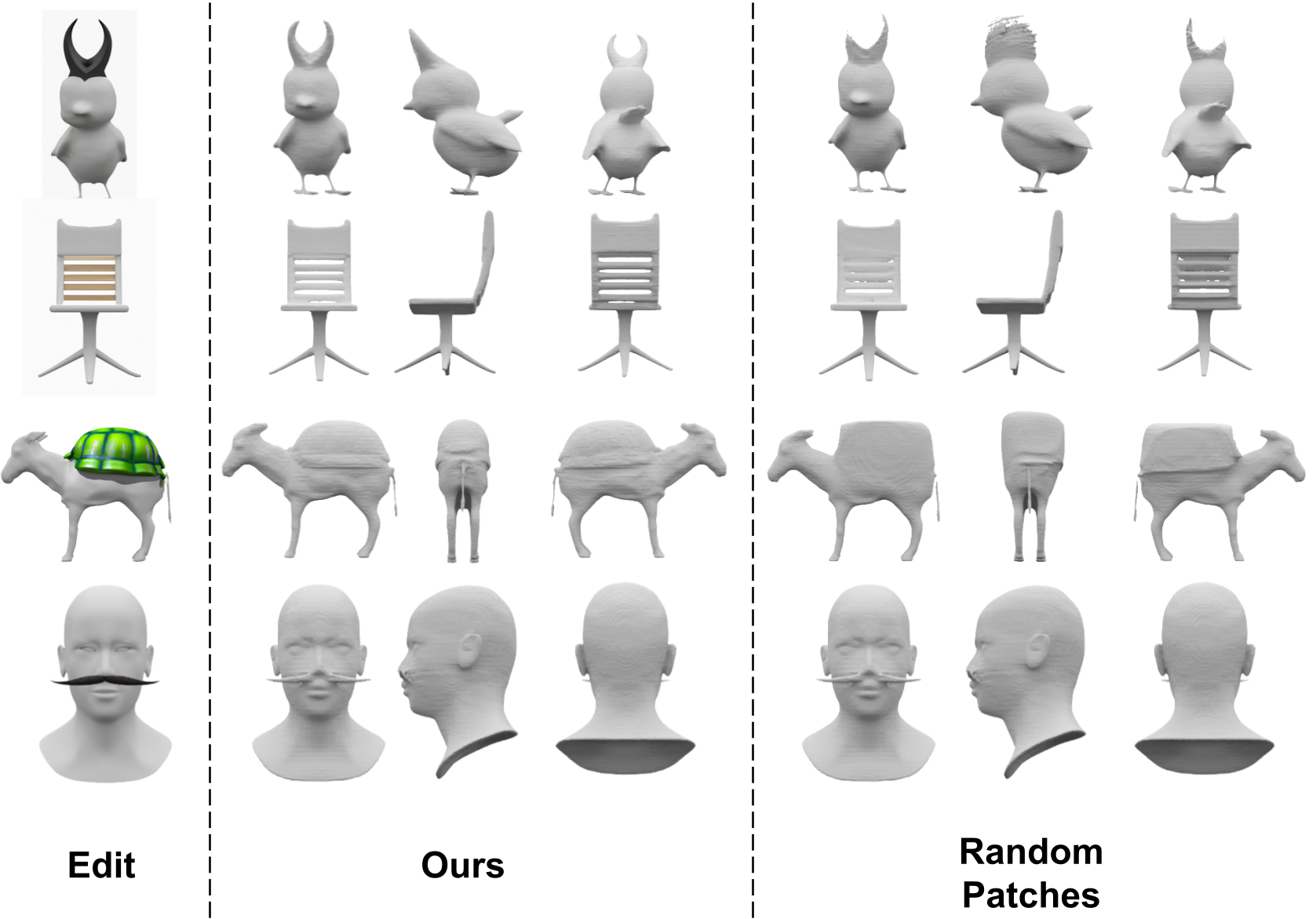}
\caption{\textbf{Impact of Random Masking}: We test our choice of masking strategy by comparing it to masking $25\%$ of patches uniformly at random. The left column shows the conditional image, the middle section shows our results, and the right section shows the results using uniformly random masking. While the model is still capable of generating correct geometry, there is a train-test gap in the masked patches since we define a contiguous 3D region to mask during inference. Thus, the model produces artifacts such as lack of sharp features (in the bird horns and chair slats) in rows 1 and 2, and overall incorrect shape in row 3 (square turtle shell).}
\label{fig:supp_ablation}
\vspace{-0.2cm}
\end{figure*}

In order to justify our masking strategy, we train our model masking patches uniformly randomly instead of using 3D occlusions. Figure~\ref{fig:supp_ablation} compares meshes extracted from a model trained using our 3D masking versus masking $25\%$ of patches uniformly at random. We observe that uniformly random patch masking can still generate ``roughly correct" shapes, especially adding a moustache to the face in the last row. This is because we add camera pose embeddings after masking, so the model can differentiate masked and non-masked tokens, regardless of their distribution within the image. Furthermore, the reconstruction outside of the masked region is still accurate. However, there still exists a train-test gap between random patches and contiguous patches created by selecting an editing region, which causes significant artifacts in the other three examples. In the first and second rows, we observe a blurring artifact, where the model cannot generate sharp features in the horns on the bird and between the slats of the chair. In the third row, using random patches causes the shape of the turtle shell to be malformed. In comparison, using our masking method produces accurate and sharp geometry in all examples.

\section*{Comparison to Tailor3D}
Tailor3D~\cite{qi2024tailor3dcustomized3dassets} is a recent work in image-to-3D generation. Similar to InstantMesh~\cite{xu2024instantmesh}, Tailor3D relies on a multiview diffusion model, namely Stable Zero-123~\cite{liu2023zero}, to generate inputs that are then lifted into 3D. Tailor3D differs in that it only requires frontal and back views, using a novel transformer design to generate 3D assets from these sparse views. 
However, Tailor3D cannot replicate our method's mesh editing results due to two sources of error. First, as with other models that rely on multi-view synthesis, inaccuracies in generating the back view propagate into the 3D model. Second, despite its unique architecture, Tailor3D still suffers from ambiguity artifacts due to the sparse input. 
Figure~\ref{fig:supp:tailor3d} demonstrates some of these artifacts. In the first two rows, we observe that Tailor3D fails to recover the geometry of the body of the bird, due to the lack of information in the front and back views, creating an incorrectly elongated shape. In the third row, we see that Stable Zero-123 completely fails to generate the back view of the wings, leading to a mirroring artifact in the final 3D shape. The fourth row suffers similar issues as the previous three, with the generated view being not only low-quality but also a mirror of the front view instead of a true back view.

\section*{Comparison to Nerfiller}
\begin{figure*}[t]
\includegraphics[width=0.98\textwidth]{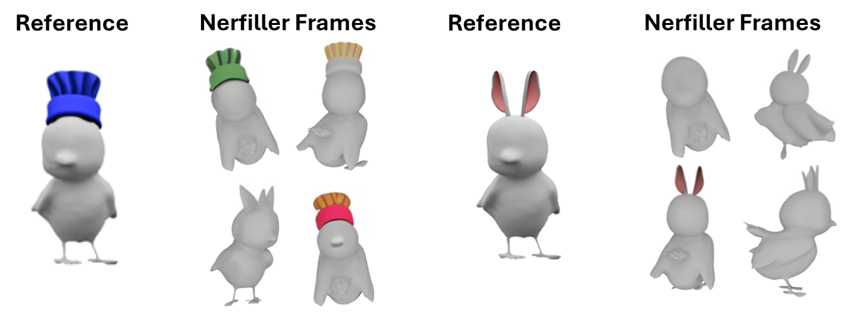}
    \caption{\textbf{Nerfiller Images}: Nerfiller results with a couple of bird edits we used for our method. We use their adapted method for reference-image based inpainting. The left column shows the reference image and the right column shows a collection of Nerfiller generated images. We observe that their inpainting method based on pre-trained diffusion models creates noisy output images. The semantics may be correct, but the details can be incorrect \eg incorrect hat colors or completely missing \eg missing hat and ears in a couple of the examples. NeRF training may be tolerant to somewhat noisy input data, but these viewpoints are not suitable for precise geometry reconstruction.}
\label{fig:supp:nerfiller}
\vspace{-0.2cm}
\end{figure*}

Nerfiller~\cite{weber2023nerfiller} is a recent work that uses pre-trained image generation models for guidance in order to inpaint masked regions in NeRFs. Nerfiller begins by training a NeRF on unoccluded pixels, and then slowly updates the training set over time via generative inpainting. They adapt their method to image-conditional completion by simply prompting the generative process using a single inpainted image as reference. This is exactly analogous to the input image edits in our method. Figure~\ref{fig:supp:nerfiller} shows some of the images Nerfiller produces using our image edits as reference. We observe that, although the inpainted images are generally semantically correct, details are inconsistent such as the color of the hats in the first row. Some frames are even missing the hat or rabbit ears entirely. While training a NeRF may tolerate some noise within the training set, this causes blurriness artifacts in the resultant 3D asset and is not suitable for explicit geometry extraction. Furthermore, since Nerfiller repeats this process of training a NeRF and then updating the dataset several times, it is significantly more expensive to run than our method, taking over an hour on an A100 GPU.

\section*{Mesh Editing Metrics}
Large scale quantitative editing comparisons are difficult as there is no standard benchmark. In the Table above we compare against MagicClay~\cite{magicclay_Barda_2024} and Instant3Dit~\cite{barda2024instant3ditmultiviewinpaintingfast}, generating edits using a text prompt. We then use the same text input for generating our conditional edited views and use CLIP similarity averaged across multiple views to measure the faithfulness of each method to the prompt.

\begin{table}[t]
\centering
\renewcommand{\arraystretch}{1.2}
\resizebox{\columnwidth}{!}{
\begin{tabular}{lcccccc} 
    \toprule
    \multicolumn{7}{c}{\textbf{Mesh Editing Comparisons}}\\
    & \multicolumn{3}{c}{ViT-L-14} & \multicolumn{3}{c}{ViT-BigG-14}\\
    \cmidrule(l){2-4}\cmidrule(l){5-7}
    & Ours & Instant3Dit & MagicClay & Ours & Instant3Dit & MagicClay \\ 
    CLIP Cos.Sim. $\uparrow$ &  0.323 & 0.303 & 0.285 &  0.337 & 0.309 & 0.286\\
    \bottomrule
\end{tabular}
}
\label{tab:mesh_editing}
\caption{\textbf{Mesh Editing Comparisons}: We provide CLIP cosine similarity metrics of our proposed MaskedLRM approach against two recent 3d editing techniques.}
\end{table}

\section*{Additional Implementation Details}
Our model implementation details are based mostly on~\cite{wei2024meshlrm}. 
Our model tokenizes \(16\times 16\) sized patches. The token embedding size and transformer width are $1024$. The transformer depth is $24$ layers. Each attention and cross attention module use multi-head attention with $16$ heads. Our model uses LayerNorm and GeLU activations with a Pre-LN architecture.

We trained our models using 64 H100 GPUs with 80GB of RAM each. We use an AdamW optimizer with $(\beta_1,\beta_2) = (0.9, 0.95)$ and a weight decay of $0.01$. During stage 1 of training, we train for $30$ epochs. For each batch consisting of $12$ shapes, we randomly sample the number of input views for the batch uniformly at random between $6$ and $8$, not including the $1$ view for the conditional view. We use another $4$ views for supervision. Over the first 1500 iterations, we linearly warm up to a peak learning rate of $4e-4$ and then use cosine learning rate decay. During stage 2, to account for increased rendering costs, we reduce the batch size to $8$ shapes. We train for $20$ epochs, with a peak learning rate of $5e-6$.

\section*{Additional Qualitative Examples}
\begin{figure*}[t]
\includegraphics[width=0.98\textwidth]{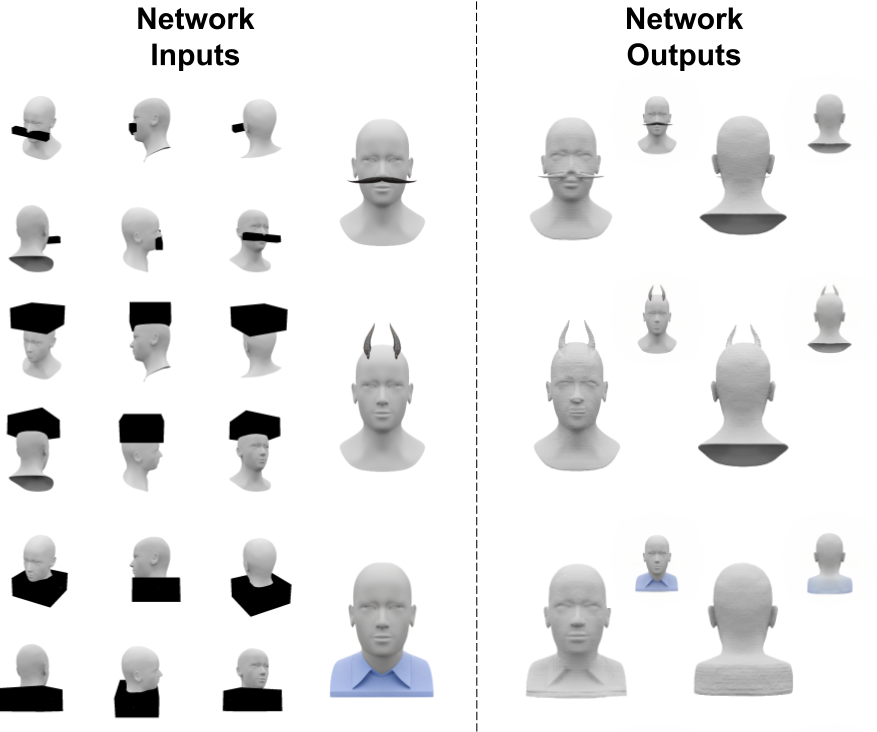}
\caption{\textbf{Additional Qualitative Examples}: Additional qualitative examples editing a person's head. The left section shows the masked views and the edited conditional image. The right section shows the mesh extracted from the network output with the volumetric renders of the SDF inset.}
\label{fig:supp:qual}
\vspace{-0.2cm}
\end{figure*}

\begin{figure*}[t]
\includegraphics[width=0.94\textwidth]{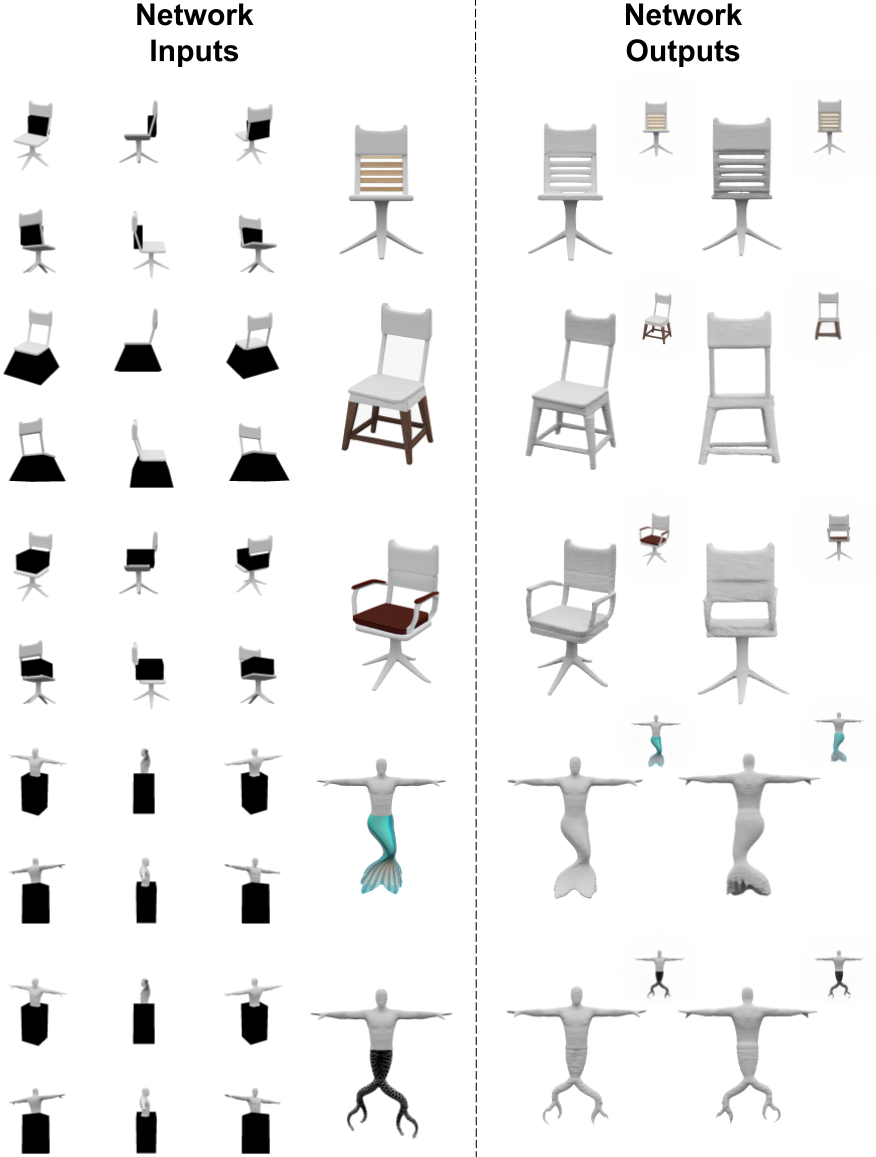}
\caption{\textbf{Additional Qualitative Examples}: Additional qualitative examples editing a chair and a full human. The left section shows the masked views and the edited conditional image. The right section shows the mesh extracted from the network output with the volumetric renders of the SDF inset.}
\label{fig:supp:qual2}
\vspace{-0.2cm}
\end{figure*}
We present some additional qualitative examples of our model in Figures~\ref{fig:supp:qual} and~\ref{fig:supp:qual2}. We show the network inputs \ie the masked views and the edited image on the left, and the network outputs \ie the resulting geometry with RGB volumetrc renders inset on the right.

\end{document}